\documentclass[10pt,journal, compsoc]{IEEEtran}
\ifCLASSOPTIONcompsoc
 \usepackage[nocompress]{cite}
\else
 \usepackage{cite}
\fi

\usepackage[switch]{lineno}

\usepackage{amsmath}
\usepackage{amsfonts}

\usepackage{etoolbox} 

\newcommand*\linenomathpatch[1]{%
  \cspreto{#1}{\linenomath}%
  \cspreto{#1*}{\linenomath}%
  \csappto{end#1}{\endlinenomath}%
  \csappto{end#1*}{\endlinenomath}%
}

\linenomathpatch{equation}
\linenomathpatch{align}

\usepackage{color}
\usepackage[normalem]{ulem}
\usepackage{subfigure}

\ifCLASSINFOpdf
\usepackage[pdftex]{graphicx}

\begin{document}

\title{Multimodal Latent Emotion Recognition from \\ Micro-expression and Physiological Signals}

\author{Liangfei~Zhang,
 Yifei~Qian,
 Ognjen~Arandjelovi\'c,
 and~Anthony~Zhu
\IEEEcompsocitemizethanks{\IEEEcompsocthanksitem L. Zhang, O. Arandjelovi\'c and A. Zhu are with the School of Computer Science, University of St Andrews, UK.\protect\\
E-mail: lz36@st-andrews.ac.uk
\IEEEcompsocthanksitem Y. Qian and A. Zhu are with the School of Mathematics and Statistics, University of St Andrews, UK.\protect\\
}}

\IEEEtitleabstractindextext{
\begin{abstract}
This paper discusses the benefits of incorporating multimodal data for improving latent emotion recognition accuracy, focusing on micro-expression (ME) and physiological signals (PS). The proposed approach presents a novel multimodal learning framework that combines ME and PS, including a 1D separable and mixable depthwise inception network, a standardised normal distribution weighted feature fusion method, and depth/physiology guided attention modules for multimodal learning. Experimental results show that the proposed approach outperforms the benchmark method, with the weighted fusion method and guided attention modules both contributing to enhanced performance.
\end{abstract}

\begin{IEEEkeywords}
Emotion recognition, multi-modal learning, micro-expression recognition, physiological signal analysis.
\end{IEEEkeywords}}

\maketitle

\section{Introduction}
Emotional states have a significant impact on physical and psychological well-being, with recognition of emotions being crucial for effective communication and understanding of individuals' emotional states and mental well-being. The complex interplay between physiology and psychology in emotional responses has led to interdisciplinary research into accurate and rapid emotion recognition, which is increasingly important in multimedia and human-computer interaction. Real-time emotion recognition has potential applications in virtual and augmented reality, healthcare, education, and marketing. In interpersonal communication, facial expressions are a critical means of conveying emotions, with \emph{micro-expressions~(MEs)} offering valuable insights into an individual's emotional state, including potential deception. MEs were initially discovered by Haggard and Isaacs in 1966 while analysing motion picture films of psychotherapy sessions for nonverbal cues between patients and therapists~\cite{Haggard1966}. Ekman and Friesen subsequently incorporated ME recognition into their deception studies~\cite{Ekman1969}, and popularised it through the TV show \emph{``Lie To Me''}. Recognising MEs enables experts to identify even the most subtle changes in an individual's facial expressions, potentially indicating their latent emotion. While facial expressions are the most reliable and universally accepted way of recognising emotions, vocal cues, body language, and physiological responses can also provide valuable information about a person's emotional state.

Enhancing emotion recognition accuracy entails exploring avenues beyond just improving the machine learning model, considering richer data types can also help achieve better performance. Human experience of the world is often multimodal, referring to how something happens or is experienced through multiple modalities. Incorporating multimodal signals can enable artificial intelligence to learn about the real world better. Relying solely on human physical signals, such as facial expression, speech, gesture, or posture, is not guaranteed as people can control these signals to hide their real emotions, especially during social communication. In contrast, \emph{physiological signals~(PS)}, which are in response to the \emph{central nervous system}~(CNS) and \emph{peripheral nervous system}~(PNS) of the human body, can provide reliable information about emotions according to Cannon's theory\cite{Cannon1927}. One significant advantage of using PS is that they are largely involuntarily activated and, therefore, difficult to control, which is a similar characteristic to ME. Researchers have attempted to establish standard relationships between emotional changes and various types of PS. 

This paper explores the benefits of incorporating multimodal data for improving latent emotion recognition accuracy, specifically with micro-expression and physiological signals. The main contributions of the present work are as follows:
\begin{itemize}[nolistsep]
    \item We introduce a novel multimodal learning framework that combines ME and PS to enhance latent emotion recognition performance. 
    
    \item We design a 1D separable and mixable depthwise inception network that effectively extracts features from various PS.
            
    \item We propose a standardised normal distribution weighted feature fusion method that reconstructs informative maps from different frames of ME video.
    
    \item We develop a guided attention module is developed that achieves multimodal learning for both micro-expression (colour and depth information) and latent emotion recognition (ME and PS).
\end{itemize}

\begin{figure*}
    \centering
    \includegraphics[width=\textwidth]{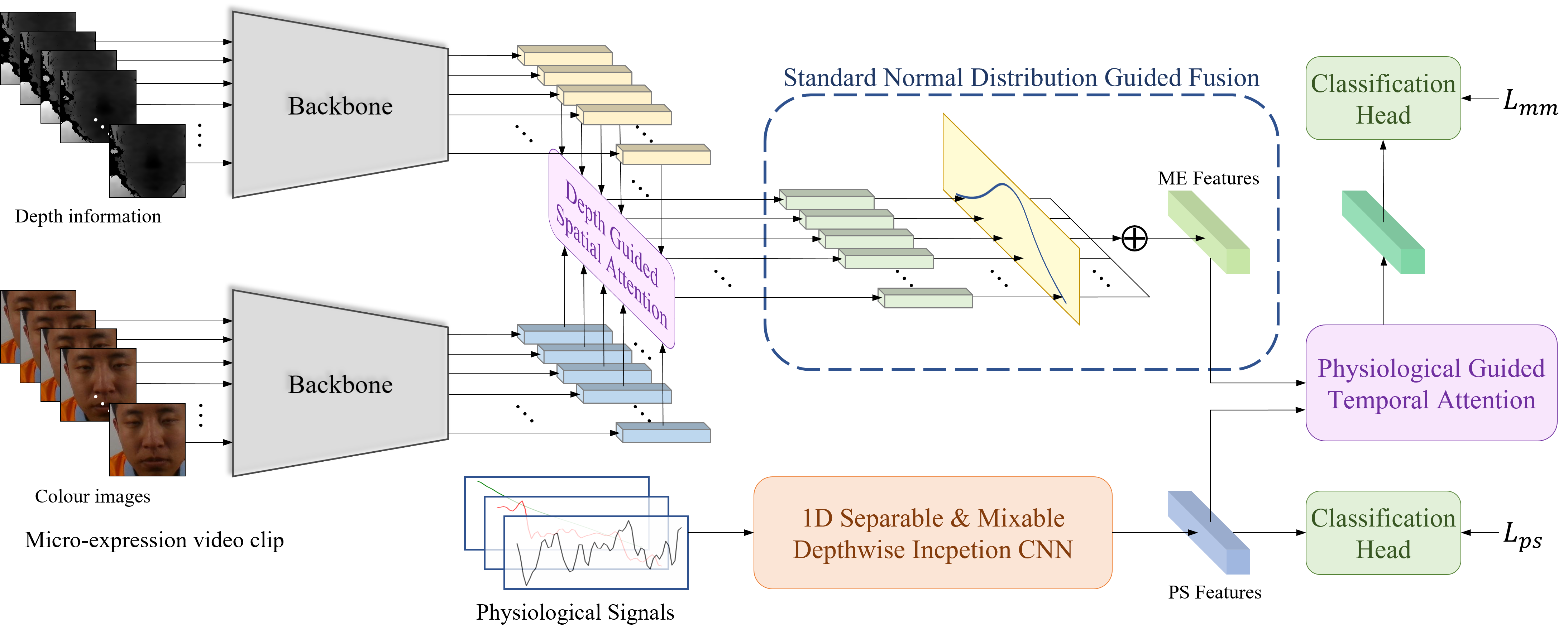}
    \caption{Architecture of the proposed framework comprising three main components: \emph{ME feature extraction branch}, \emph{PS feature extraction branch}, and \emph{guided attention fusion module}. The final loss is $(L_{ps} + L_{mm})/2$, where $L_{ps}$ and $L_{mm}$ are both cross-entropy losses calculated from the PS branch and whole multimodal learning, respectively. 
    }
    \label{fig:framework}
\end{figure*}

\section{Related Work}
\label{sec:related_work}
\subsection{Micro-expression recognition}
Micro-expression recognition~(MER) which refers to the recognition of emotions expressed in a sequence of faces known to be brief and subtle, is one significant challenge in affective computing to mining human's authentic emotions. In recent years, computer vision technology has been increasingly utilised for automatic MER, which has improved the feasibility of applications involving MEs~\cite{Zhang2021}. Since the publication of the open spantenous ME databases~\cite{Li2013,Yan2013} in 2013, the volume of research on automatic MER has been increasing steadily over the years. From the handcrafted computer vision methods in the early years to the deep learning approaches more recently, the main ideas of ME feature extraction could be categorised as primarily pursuing either a spatial strategy or a temporal one. 

Facial Region Of Interest~(ROI) segmentation is a common technique for spatial feature extraction in ME analysis~\cite{Zhang2021a,Wang2014b,Liu2016}, where the face sample is segmented into different regions based on the Facial Action Coding System~(FACS)~\cite{Ekman2002}. These regions can correspond to independent facial muscle complexes and are subjected to individual appearance normalisation. Polikovsky~et al. introduced a gradient feature that constructs histograms of local gradient projections across different regions~\cite{Polikovsky2009}, and the Local Binary Pattern (LBP) operator describes local appearance in a robust way using the relative brightness of neighbouring pixels~\cite{Pfister2011}. As for deep learning approaches, the convolution and pooling layers perform spatial feature extraction. Attention blocks have been introduced in recent years to improve the learning of spatial features within neural networks, by generating weight masks for feature maps to focus on significant regions. Graph Convolutional Network~(GCN) are another optimisation measure used to capture spatial information, often by using Action Units~(AUs) as corresponding graph nodes~\cite{Xie2020}. These methods use a priori knowledge to enhance the extracted spatial features.

The sudden appearance of MEs makes their temporal features important. While some methods use only the apex frame of each ME sample~\cite{Peng2019a,Gan2019,Liong2018}, most use all frames between the onset and offset, treating all temporal changes equally. Some even use temporal frame interpolation to increase the number of frames~\cite{Li2018,Liu2016,Khor2018,Wang2014b,Pfister2011}. Many handcrafted feature-based approaches treat raw video data as a 3D spatio-temporal volume, applying the same operator used to extract spatial features to pseudo-images formed by a cut through the 3D volume comprising one spatial dimension and the temporal dimension. LBP-TOP and 3DHOG are examples of this approach. Similar in this regard are optical flow-based features, which combine local spatial and temporal elements~\cite{Liong2019,Liong2018,Liu2016}. Instead of using raw appearance imagery, some authors propose using pre-processed data in the form of optic flow matrices to exploit proximal temporal information directly in deep learning methods~\cite{Xia2020,Liu2019,Kumar2021,Zhang2022}. Longer-range temporal patterns are learned by treating video sequences as 3-dimensional matrices or by employing structures like RNN or LSTM~\cite{Khor2018,Kim2016a}. 

In previous research, colour images or videos have been the primary data samples used due to the availability of ME databases. However, the recent release of 4DME~\cite{Li2022} and CAS(ME)$^3$~\cite{Li2023} has expanded the range of data available for emotion recognition related to micro-expressions. 4DME primarily focuses on 3D micro-expression data, while CAS(ME)$^3$ not only includes RGB-D micro-expression video clips but also includes physiological signals in one part of the database. This enables multimodal learning with micro-expressions for emotion recognition.

\subsection{Multimodal emotion recognition}
Numerous works in the literature show that multimodal emotion recognition systems outperform unimodal approaches~\cite{D2015,Nahid2021}. Emotions are intricate experiences that involve not only observable physical expressions but also internal feelings, thoughts, and other processes that may not be consciously perceived by the individual. For instance, a person may force a smile during a formal social occasion even if they are experiencing negative emotions. The other approach to detect emotions is through PS such as electroencephalogram~(EEG), temperature, electrocardiogram~(ECG), electromyogram~(EMG), galvanic skin response~(GSR), respiration~(RSP), etc. ECG is a non-invasive method for measuring the electrical activity of the heart, which can be displayed as a waveform on a computer screen or chart recorder. This helps in identifying whether a heartbeat is normal or abnormal. While EEG signals are commonly used for emotional expression recognition due to their robust sensing of emotions in the brain~\cite{Munoz2018,Nakisa2018,Pandey2019}, recent studies have explored the use of ECG-based recognition~\cite{Bras2018, Kaji2018}, though the number of such studies is limited. The high dimensionality of EEG signals makes it difficult to identify effective features for emotional expression recognition, and thus some techniques have been proposed to fuse several PS for emotion detection. 

The fusion module is the most crucial aspect of multimodal learning. The techniques include early fusion, which involves combining extracted features from signals before sending them to the classifier, and late fusion, which involves taking the final result by voting the results produced by several classifiers. The most straightforward approach in early fusion is to concatenate the feature vectors from all modalities, also known as plain early fusion. In the work by Verma and Tiwary, plain early fusion was employed to fuse energy-based features extracted from 32-channel EEG signals~\cite{Verma2014}. Another way to encode feature dependencies is to use probabilistic inference models like Hidden Markov Models and Bayesian Networks~(BNs), for example by building a BN to fuse features from both EEG and ECG signals~\cite{Shin2017}. Late fusion, on the other hand, uses multiple classifiers that can be trained independently, and the final decision is made by combining the outputs of each classifier. A framework for emotion recognition based on the weighted fusion of basic classifiers was proposed~\cite{Wang2015c}, where three support vector machines using different features were developed and their results were combined using weighted fusion based on each classifier's confidence estimation for each class. 

Several studies have attempted to combine facial expressions with physiological data to create accurate emotion recognition systems~\cite{Cimtay2020,Huang2017}. Both studies used late fusion techniques, such as voting and decision tree, to combine the decisions made from facial expressions and PS, though Cimtay~et al. use early fusion for different signals. However, people may conceal their true emotions behind fake facial expressions, whereas MEs can reveal their genuine emotions, as can PS. Therefore, combining MEs and PS through multimodal learning could be a better strategy for authentic emotion recognition. Li~et al. attempted to use voice, electrodermal activity~(EDA) and depth information to assist MER. They converted the voice and EDA signals into 2D greyscale input channels and trained them with colour and depth information from the apex frame~\cite{Li2023}. However, their results of combining EDA or speech signals were not satisfactory due to not addressing the noise in the signals or designing a specific network for PS. Despite this, the database they provided is a valuable resource for researchers to optimize multimodal emotion recognition processing with MEs.

\section{Proposed method}
Colour images are a crucial and widely used source for computer vision tasks~\cite{Qian2022,Shu2022,Conti2022}, providing valuable information for deep learning models to analyse and interpret visual data. In addition, ME has been extensively studied and recognised as a valuable source for authentic emotion recognition, as discussed in Section~\ref{sec:related_work}. Therefore, in our proposed framework, we consider colour images from ME video clips as the primary source, with depth information used to guide the spatial features from each frame. Features are extracted from colour images and depth maps separately using backbone networks. To fuse the spatial features extracted from each frame, we designed a standard normal distribution guided fusion method that pays more attention to the middle, where facial movements usually reach their apex, and less attention to the ends. Apart from ME, PS are used in another branch of the proposed framework to recognise latent emotions by our designed 1D separable and mixable depthwise inception network and enhance spatiotemporal features extracted from the ME sample. By analysing both ME and PS, the network can gain a deeper understanding of the subject's emotional state and achieve more accurate latent emotion recognition results. Figure~\ref{fig:framework} shows the proposed framework. 

\subsection{1D separable \& mixable depthwise inception CNN}

\begin{figure*}
    \centering
    \includegraphics[width=\textwidth]{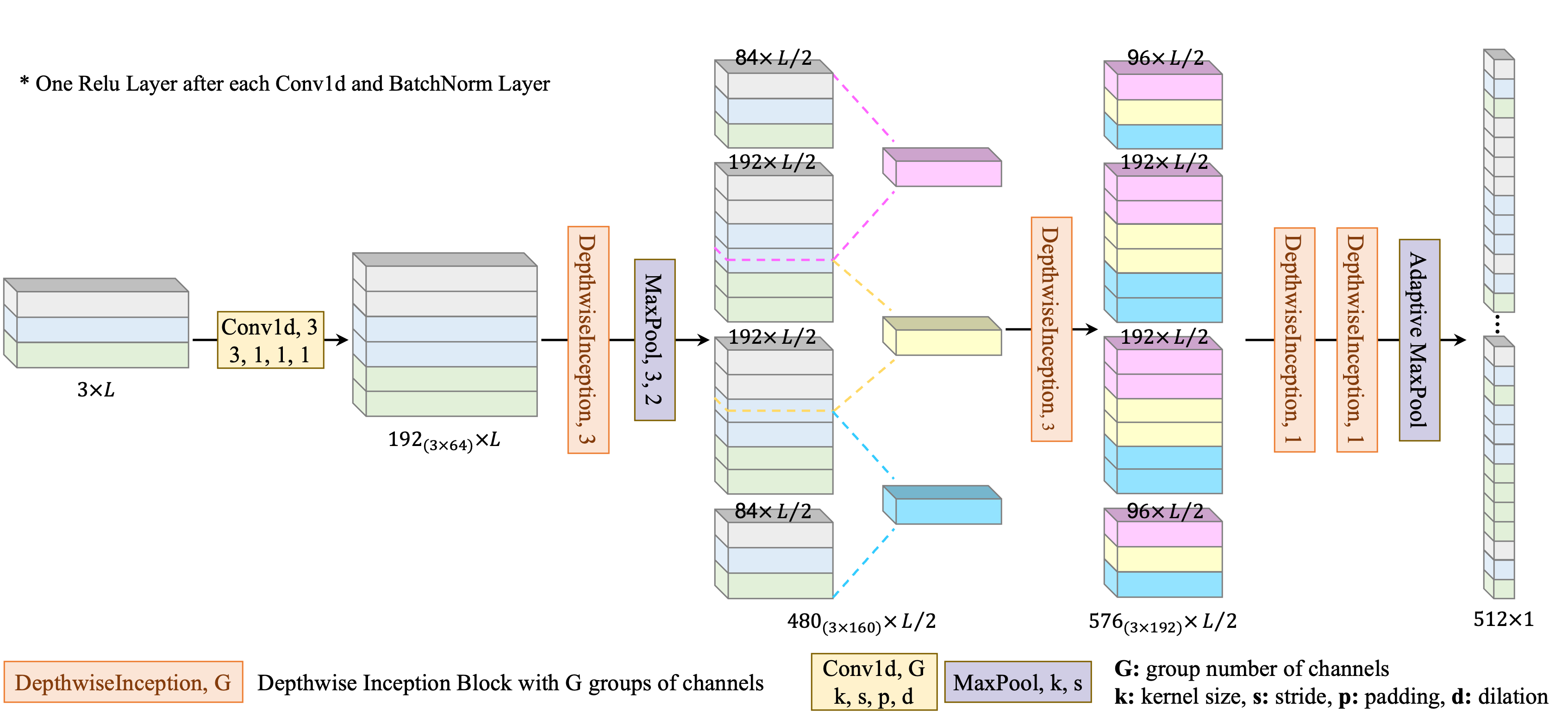}
    \caption{The separable and mixable network proposed for physiological signals, see Figure~\ref{fig:DIblock} for details of depthwise inception block, where $L$ is the 1D input length of signals.}
    \label{fig:1DSMDIN}
\end{figure*}

\begin{figure}
    \centering
    \includegraphics[width=0.3\textwidth]{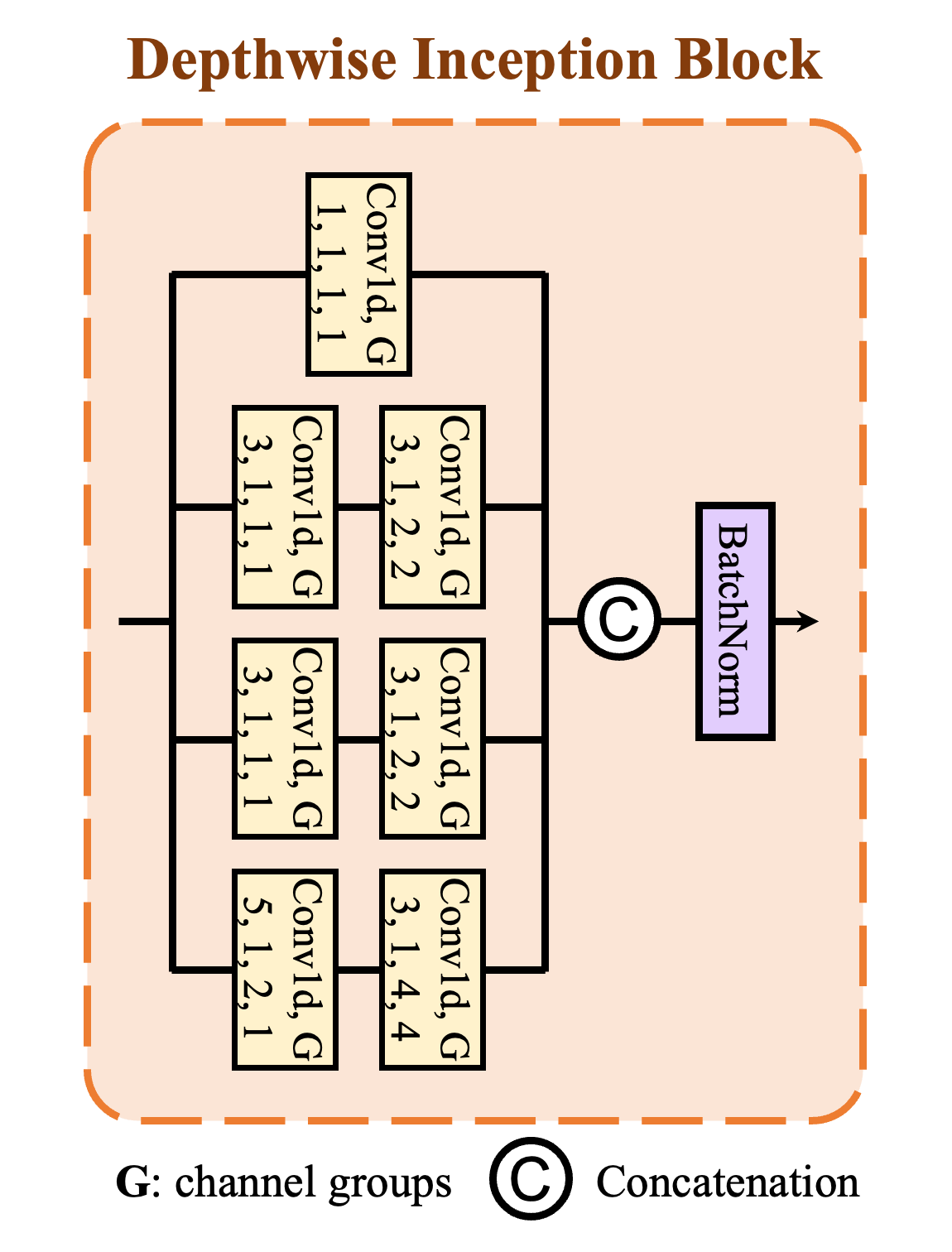}
    \caption{Illustration of the depthwise inception block's network design and layer hierarchy.}
    \label{fig:DIblock}
\end{figure}

The network is designed to extract PS features in our framework, whose structure can be seen as Figure~\ref{fig:1DSMDIN}. The depthwise structure contains separate convolutions for each group of channels, allowing for more precise feature extraction and capturing spatial correlations. This enables the network to learn more complex and diverse representations of the input data. The Inception block uses a combination of convolutional filters of different sizes within a single layer. This allows the network to capture features at multiple scales without the need for multiple layers, which can be computationally expensive. The Inception block can be seen as an ensemble of smaller networks with different filter sizes, which provides a form of regularisation that helps prevent overfitting and improves generalisation performance.

In order to effectively extract features from multiple input channels, a method is proposed that involves the extraction of features from each channel individually, followed by their combination for further learning. To achieve this, a depthwise convolutional layer is initially employed to enhance the features from each input channel. Subsequently, a depthwise inception block with three groups is designed to extract features from each physiological source. The design of the depthwise inception block is illustrated in Figure~\ref{fig:DIblock}, and comprises four branches of depthwise convolutional layers with varying kernel sizes to extract features from different scales. The resulting features from each branch are then concatenated together and fed into the next depthwise inception block. Since the number of output channels in each branch of the inception block is different, the features from different sources are mixed together to form a new group of channels for further learning. Finally, the mixed features are fed into the last two blocks as a whole group to extract the final features from the physiological sources.

\subsection{Standardised normal distribution weighted feature fusion}

The sequence of extracted features is- mapped to the ME feature with a standard normal distribution function. Note that the standard normal distribution is the special case of the normal distribution with mean \(\mu = 0\) and variance \(\sigma^2 = 1\). A technicality is that since the frames are discrete, the function is slightly altered mapping from a set of extracted features of all the frames to a set of values within the range (0, 1) representing the weight of features from the set. It is generally believed that, during the instance of ME, the most prominent facial movement is roughly in the middle of the timeframe, namely, the apex frame of a ME sample is roughly in the middle of the clip. In addition, the frames closer to the apex frame contains more valuable features than those more distant from the apex frame. In contrast to extracting spatial features only in the apex frame, we fuse all the features that are laid in several adjacent frames, i.e. the whole duration. Instead of a uniformly weighted function where each feature is the same weight, in our approach, features extracted from frames which are closer to the middle of the clip are considered more representative and valuable and thus are mapped to higher weight. This is in line with the proposition stated above. The function is as follows:
\begin{equation}
    W_f = \frac{\exp\left(-\frac{i^2}{2}\right)}{\sum_{f=0}^{F-1} \exp\left(-\frac{i^2}{2}\right)}, \quad\text{where}\quad
    i = -3 \sigma + f \cdot \frac{6 \sigma}{F-1},
\end{equation}
where $F$ denotes the number of frames in one ME sample. Note that roughly 99.7\% of the probability density for a standard normal distribution lies within $3$ standard deviations of the mean, and thus lies within the range $(-3, 3)$. Only the weights within $3$ standard deviations are considered and the remaining 0.3\% is negligible.

\subsection{The depth/physiology guided attention module}

\begin{figure}
    \centering
    \includegraphics[width=0.48\textwidth]{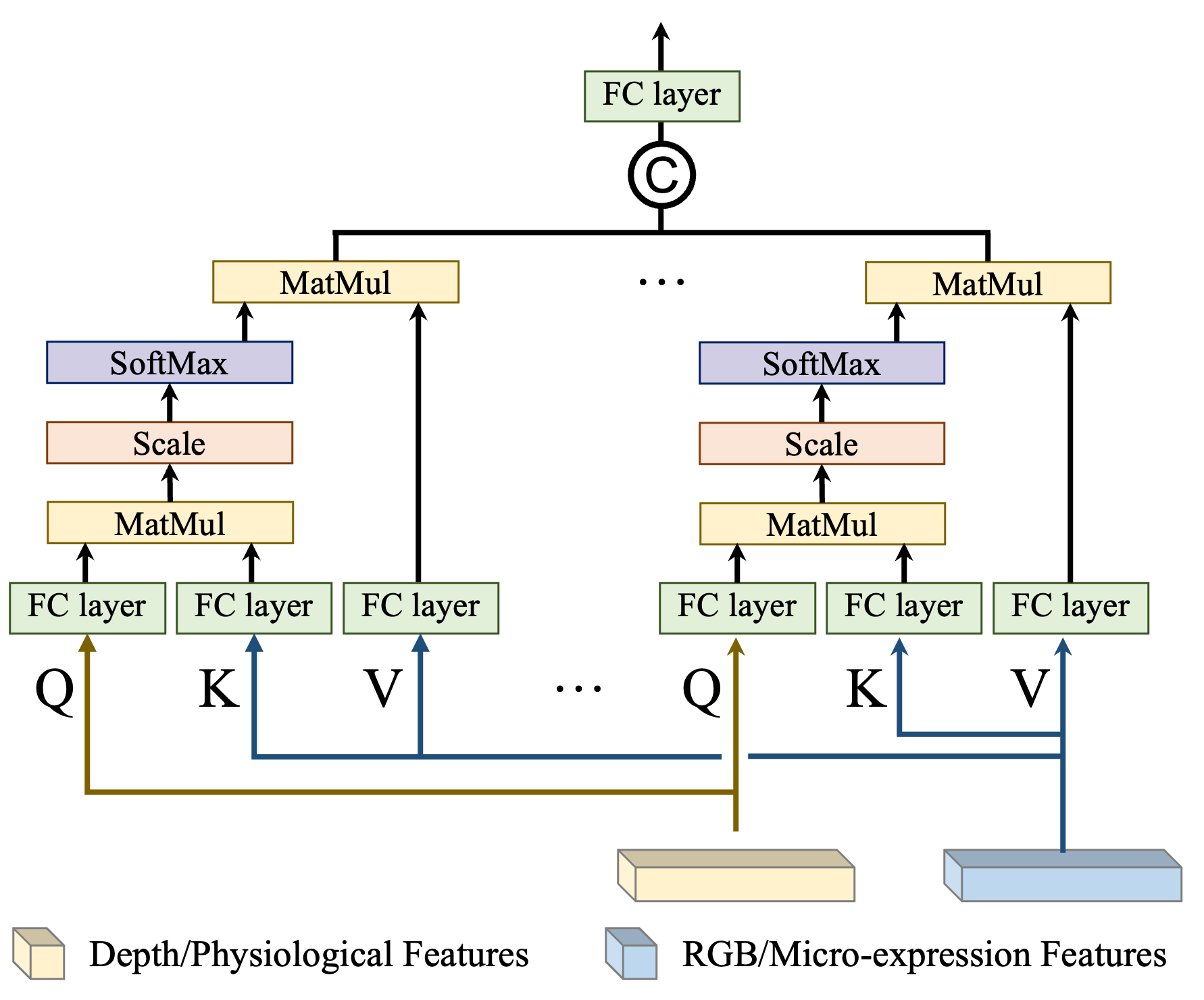}
    \caption{Structure of the guided attention module.}
    \label{fig:attention}
\end{figure}

The module is designed for feature fusion of both colour and depth information for each frame of micro-expression, as well as the final fusion of ME and PS features. For attention modules, formally we have a query $Q$, a key $K$, and a value $V$ to calculate attention. The depth and PS features could be considered as the input $Q$ to guide attention learning. At the beginning of the module, the main features are copied as sources for both inputs $K$ and $V$. After fully-connected layers, the scaled dot-product attention mechanism, denoted as $\text{SDP}$ below, is run through several times in parallel. The scaled dot-product attention is an attention mechanism where the dot products are scaled down by $\sqrt{d}$: 
\begin{equation}
\text{SDP}(Q, K, V) = \text{softmax}\left(\frac{QK^{T}}{\sqrt{d}}\right)V.
\end{equation}
$d$ is the dimension of the queries and keys, and $\text{softmax}$ denotes the softmax function. The dot product results of the attention mechanism are divided by $\sqrt{d}$ to maintain a variance of $1$. The independent attention outputs are then concatenated and linearly transformed to the expected dimension. The multi-head attention mechanism is defined as follows:\\[-3pt]
\begin{equation}
\begin{aligned}
 \text{MultiHead}&\left(\textbf{Q}, \textbf{K}, \textbf{V}\right) = \left[\text{head}_{1},\dots,\text{head}_{h}\right]\textbf{W}_{0}, \\
 \text{head}_{i} &= \text{SDP} \left(\textbf{Q}\textbf{W}_{i}^{Q}, \textbf{K}\textbf{W}_{i}^{K}, \textbf{V}\textbf{W}_{i}^{V} \right), 
\end{aligned}
\end{equation}
where $\textbf{W}$ represents the learnable parameter matrices. The multi-head attention mechanism allows for different parts of the sequence to be attended to differently, such as longer-term dependencies versus shorter-term dependencies.

\begin{figure*}
    \centering
    \includegraphics[width=\textwidth]{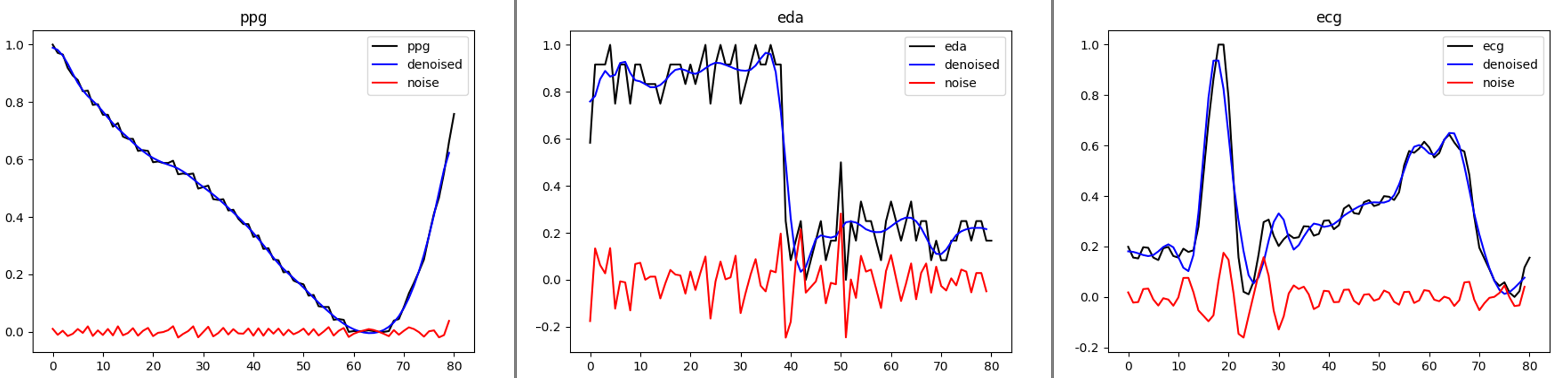}
    \caption{Examples of Daubechies wavelet denoising results for physiological signals.}
    \label{fig:sig_denoise}
\end{figure*}

\section{Experiments and evaluation}

\subsection{Data preparation and experiment setting}
We use Part~C of CAS(ME)$^3$ corpus~\cite{Li2023} developed to address the challenges of ME elicitation, collection, and annotation. CAS(ME)$^3$ is composed of three parts. Part~A and~B contain labelled and unlabelled long videos recorded in the same environment and labelled by the same labellers. CAS(ME)$^3$ uniquely introduces multimodality to ME analysis in Part~C, which is a third-generation multimodal spontaneous ME database that goes beyond just RGB images and includes depth information, voice, and PS. Part~C used a third-generation of ME eliciting paradigm, mock crime, with higher ecological validity. Participants were asked to steal a small amount of money from an envelope and were subsequently questioned about the theft. The scenario was designed to create a stressful situation that would elicit spontaneous MEs associated with guilt or deception. Part~C contains 166 MEs from 31 subjects and makes it possible to enrich multimodal ME analysis with PS, including electrodermal activity~(EDA), heart rate/fingertip pulse -- electrocardiogram~(ECG), respiration~(RSP), and pulse photoplethysmography~(PPG). 

The colour and depth frames in Part~C are captured at a frame rate of 30~fps. Due to the definition of the happening time of a micro-expression, we selected only the samples with less than 15 frames (500ms) from the database. We cropped the facial region based on the landmarks detected from the onset colour image, and this cropping was applied to all subsequent colour and depth images. To process the data, we utilised pretrained VGG-Face~\cite{Parkhi15} and VGG-16~\cite{Simonyan2014} networks as backbone networks for colour and depth, respectively.

As for the PS branch, we utilised EDA, ECG, and PPG signals as the three-channel input. To process the source signal data, we employed wavelet denoising on the segmented signal clips. This method is highly effective in denoising 1D signals due to its ability to capture both local and global features of the signal accurately, whilst maintaining a good balance between time and frequency localisation. Daubechies wavelets are orthogonal and form a complete basis set, allowing the signal to be decomposed into its wavelet coefficients, which can then be thresholded to remove noise, as shown in Figure~\ref{fig:sig_denoise}.

\subsection{Experimental results}
The traditional evaluation approach for micro-expression recognition involves leave-one-subject-out (LOSO) cross-validation, where a single subject's data is withheld and used as a validation data set, while all remaining subjects' data is used for training. The overall performance of a method is then assessed by aggregating the results of all different possible iterations of the process, i.e.\ of all subjects being withheld in turn. Li~et al. also utilised LOSO in their experiments~\cite{Li2023}. To ensure fair comparison and more accurate evaluation, we applied the LOSO approach in our experiments as well. Accuracy, unweighted F1-score (UF1) and unweighted average recall (UAR), averaging the per-class recall and F1-score respectively, are used as metrics during evaluation.

Our study aimed to investigate multimodal latent emotion recognition, and the main results can be seen in Table~\ref{tab:results}. To confirm the effectiveness of our proposed network structure and each designed module, we conducted ablation experiments. Table~\ref{tab:distribution} presents the results related to the standardized normal distribution guided spatial feature fusion, while Table~\ref{tab:attention} displays the results of the depth/physiology guided attention module.

\begin{table*}
    \centering
    \caption{Comparison of multimodal analysis for latent emotion recognition. ``Colour'' and ``Depth'' are from micro-expression samples, and ``PS'' indicates the combination of EDA, PPG, and ECG in our results, while representing the use of only EDA in Li~et al.'s results.}
    \label{tab:results}
    \resizebox{\textwidth}{!}{
    \begin{tabular}{|l|c|c|c|c|c|c|c|c|c|c|c|c|}
    \hline & \multicolumn{3}{c|}{ \textbf{Colour} } & \multicolumn{3}{c|}{ \textbf{Colour + Depth} } & \multicolumn{3}{c|}{ \textbf{Colour + Depth + PS} } & \multicolumn{3}{c|}{ \textbf{Colour + PS} } \\
    \hline & Acc & UF1 & UAR & Acc & UF1 & UAR & Acc & UF1 & UAR & Acc & UF1 & UAR \\
    \hline \textbf{Li~et al.\cite{Li2023}} & - & 0.248 & 0.263 & - & 0.296 & 0.296 & - & 0.230 & 0.244 & - & 0.230 & 0.260 \\
    \hline \textbf{Ours} & $\mathbf{0 . 6 4 0}$ & $\mathbf{0 . 3 5 3}$ & $\mathbf{0 . 3 4 5}$ & $\mathbf{0 . 6 4 0}$ & $\mathbf{0 . 3 1 5}$ & $\mathbf{0 . 3 1 8}$ & $\mathbf{0 . 7 3 8}$ & $\mathbf{0 . 5 8 6}$ & $\mathbf{0 . 5 6 3}$ & $\underline{\mathbf{0 . 7 5 0}}$ & $\underline{\mathbf{0 . 6 4 2}}$ & $\underline{\mathbf{0 . 5 7 8}}$ \\
    \hline
    \end{tabular}}
\end{table*}

\begin{table*}
    \begin{minipage}{0.43\textwidth}  
    \centering
    \caption{Results of the comparison study on standard normal distribution fusion for ME recognition.}
    \resizebox{\columnwidth}{!}{
    \label{tab:distribution}
    \begin{tabular}{|l|c|c|c|}
    \hline & \multicolumn{3}{c|}{ \textbf{Colour} } \\
    \hline & Acc & UF1 & UAR \\
    \hline \textbf{Uniform Distribution} & 0.610 & 0.258 & 0.283 \\
    \hline \textbf{Standard Normal Distribution} & $\mathbf{0 . 6 4 0}$ & $\mathbf{0 . 3 5 3}$ & $\mathbf{0 . 3 4 5}$ \\
    \hline
    \end{tabular} }
    \end{minipage}
    \begin{minipage}{0.57\textwidth}
    \centering
    \caption{Results of the comparison study on the impact of depth and spatial guided attention modules for multimodal latent emotion learning.}
    \resizebox{\columnwidth}{!}{
    \label{tab:attention}
    \begin{tabular}{|l|c|c|c|c|c|c|}
    \hline & \multicolumn{3}{c|}{ \textbf{Colour + Depth} } & \multicolumn{3}{c|}{ \textbf{ME + PS} } \\
    \hline & Acc & UF1 & UAR & Acc & UF1 & UAR \\
    \hline \textbf{Concatenation} & 0.604 & 0.262 & 0.288 & 0.701 & 0.492 & 0.468 \\
    \hline \textbf{Guided Attention module} & $\mathbf{0 . 6 4 0}$ & $\mathbf{0 . 3 1 5}$ & $\mathbf{0 . 3 1 8}$ & $\mathbf{0 . 7 3 8}$ & $\mathbf{0 . 5 8 6}$ & $\mathbf{0 . 5 6 3}$ \\
    \hline
    \end{tabular} }
    \end{minipage}
\end{table*}

\subsection{Analysis and discussion}
All results of our experiments will be discussed in this section. We start by investigating the impact of multimodal learning on latent emotion recognition performance. Then, we discuss the effectiveness of each design inside the proposed framework.

The results presented in Table~\ref{tab:results} demonstrate a significant improvement in performance compared to the benchmark results of Li~et al. who used RGB and depth information from the apex frame only as 4-channel input for AlexNet. While producing worse performance than the proposed approach their work demonstrated the value of depth information. In contrast to Li~et al., herein we used all frames from a video clip and a standard normal distribution feature fusion module to merge the features extracted from all frames. We note that although the use of depth information facilitates the learning of more expressive features, it may introduce noise, which is particularly problematic in micro-expression analysis wherein the signal corrupted by noise is weak; therefore, one of the potential avenues for further research in colour and depth micro-expression recognition could be finding a better approach to denoise the depth information. Regarding physiological signals, Li~et al. used them as greyscale 2D input channels to the same backbone, without addressing their noise content or designing a specialized network to process them. In contrast, the proposed approach employed the Daubechies wavelet for denoising and introduced a 1D separable and mixable depthwise inception CNN for feature extraction. Our results suggest that this network structure contributes to the improved performance in recognising latent emotions.

The performance of the proposed standardised normal distribution weighted fusion method is compared with that of the uniform distributed fusion method in Table~\ref{tab:distribution}. The results demonstrate that the proposed method can fuse the features extracted from each frame of micro-expressions more effectively than simply adding them together. The weighted fusion method assigns different weights to different features based on their importance, allowing more important features to have a greater influence on the overall learning process. This emphasises the significance of each feature's contribution to the final result and enhances the performance of the model in recognising micro-expressions. Furthermore, a comparison study was conducted to evaluate the impact of depth/physiology guided attention modules on the performance of the proposed model, a comparison study was conducted. We used concatenation as the baseline fusion method for multimodal learning and trained and tested four different configurations of the model. The results of the study, presented in Table~\ref{tab:attention}, revealed that the depth-guided attention module outperformed concatenation in incorporating colour and depth information. Additionally, the physiology-guided attention module used for emotion recognition demonstrated significantly better results, indicating that these guided attention modules are capable of effectively fusing extracted features from multiple modalities to learn more beneficial mixed features and contribute to the improved performance of the proposed model in latent emotion recognition.

\section{Conclusion}
Emotional states have a significant impact on physical and psychological well-being, and the recognition of emotions is essential for effective communication and understanding of an individual's emotional and mental state. However, relying solely on facial expressions is not sufficient as people can control these signals to hide their real emotions, especially during social communication. Therefore, this paper proposes a multimodal learning framework that combines micro-expressions and physiological signals to enhance latent emotion recognition performance. The proposed approach denoises the signals and uses a 1D separable and mixable depthwise inception CNN for physiological feature extraction. Furthermore, the paper proposes a standardised normal distribution weighted feature fusion method and a guided attention module that achieves multimodal learning for both micro-expression and latent emotion recognition. The results show a significant improvement in performance compared to the benchmark results, demonstrating the potential benefits of incorporating multimodal data for improving latent emotion recognition.

\bibliographystyle{IEEEtran}

\end{document}